%% file: arXiv_treed_bo.tex
\documentclass[twoside]{article}
\usepackage{arXiv}

\usepackage{times}

\usepackage{hyperref}
\usepackage{url}

\usepackage{amssymb}
\usepackage{amsmath,amsthm,amssymb,amsfonts,amsthm}
\usepackage{dsfont}
\usepackage{graphicx}
\usepackage{wrapfig}
\usepackage[font={small}]{caption}
\usepackage[font={small}]{subcaption}

\input{include/murphy}

\DeclareMathOperator*{\argmax}{arg\,max}
\DeclareMathOperator*{\T}{^\intercal}

\usepackage[]{natbib}

\begin{document}

\twocolumn[
\aistatstitle{Heteroscedastic Treed Bayesian Optimisation}
\aistatsauthor{\hspace{0.2in}John-Alexander M. Assael \And Ziyu Wang \And Bobak Shahriari \And Nando de Freitas}
\aistatsaddress{\hspace{0.2in}Imperial College London \And University of Oxford \And University of British Columbia \And University of Oxford\\ CIFAR Fellow}
]

\begin{abstract}
Optimising black-box functions is important in many disciplines, such as tuning machine learning models, robotics, finance and mining exploration.
Bayesian optimisation is a state-of-the-art technique for the global optimisation of black-box functions which are expensive to evaluate.
At the core of this approach is a Gaussian process prior that captures our belief about the distribution over functions. 
However, in many cases a single Gaussian process is not flexible enough to capture non-stationarity in the objective function. Consequently, heteroscedasticity negatively affects performance of traditional Bayesian methods.
In this paper, we propose a novel prior model with hierarchical parameter learning that tackles the problem of non-stationarity in Bayesian optimisation. 
Our results demonstrate substantial improvements in a wide range of applications, including automatic machine learning and mining exploration.
\end{abstract}

\section{Introduction}

Bayesian optimisation has proven to be a popular and successful methodology for global optimisation of expensive, black-box functions. It is used to find the global minimum of generally non-convex, multi-modal functions whose derivatives are unavailable. The evaluations of the objective function are often only available via noisy observations. Major applications of these techniques include interactive user interfaces \citep{Brochu:2010} robotics \citep{Lizotte:2008, Martinez-Cantin:2009}, environmental monitoring \citep{marchant2012bayesian}, estimating thermophysical properties of materials \citep{assael2014novelportable}, information extraction \citep{Wang:2014hyper}, sensor networks \citep{Garnett:2010, Srinivas:2010}, adaptive Monte Carlo \citep{Wang:ahmc}, experimental design \citep{Azimi:2012}, and reinforcement learning \citep{Brochu:2009}. An application that has inspired great interest recently is that of automatically tuning machine learning algorithms \citep{Hutter:2011, Bergstra:2011, Snoek:2012, Swersky:2013, Thornton:2013, Hoffman:2014}.

In general, the goal of global optimisation is to find the optimum
\begin{equation}
\vx^* = \argmax_{\vx \in \calX} f(\vx),
\end{equation}
of an objective function $f: \calX \mapsto \mathbb{R}$ over an index set $\calX \subset \real^d$. The approach of Bayesian optimisation may be understood in the setting of sequential decision making, whereby at the $t$-th decision round, we select an input $\vx_t \in \calX$ and observe the value of the {black-box} reward function $f(\vx_t)$. The returned value $y_t$ may be deterministic, $y_t = f(\vx_t)$, or stochastic, $y_t = f(\vx_t) + \epsilon_t$, where $\epsilon_t$ is a noise process.

Since the function is unknown, we use a Bayesian prior model to encode our beliefs about its smoothness, and an observation model to describe the data $\data_t=\{(\vx_i, \vy_i)\}_{i\leq t}$ up to the $t$-th round. Using these two models and the rules of probability, we derive a posterior distribution $p(f|\data_t)$ that can in turn be used to build an acquisition function to decide the next input query $\vx_{t+1}$. The acquisition function trades-off exploitation and exploration in the search process.
For a comprehensive introduction of Bayesian optimisation, please refer to~\citep{Brochu:2009,Snoek:2012}.

Most of the aforementioned practical applications tend to have heteroscedastic objective functions.
\cite{snoek:2013b} addressed this fundamental problem using warped Gaussian processes. In this work, we introduce a flexible novel model for dealing with heteroscedasticity. In particular, we adopt trees with (warped) Gaussian process leaves. We explain how to construct these trees properly avoiding variance explosion near split points. We also introduce a hierarchical approach for learning the hyper-parameters, so as to address situations in which only a few points are observed at each leaf. All these methodological improvements, where combined resulting significantly improved empirical performance in a wide range of applications.

\section{Background}
In this section, we give a brief overview of Bayesian optimisation 
as well as a brief survey on Heteroscedastic Gaussian processes.

\subsection{Gaussian processes}
Gaussian processes (GPs) are popular priors for Bayesian optimisation as they offer
a simple and flexible model to capture our beliefs about the behaviour of the
function;
we refer the reader to~\cite{Rasmussen:2006} for details on these stochastic
processes. These priors are defined by a mean function $m:\calX\mapsto\real$ and a covariance kernel $k:\calX\times\calX\mapsto\real$. Given any collection of inputs $\vx_{1:t}$, the outputs are jointly Gaussian:
\begin{equation*}
    f(\vx_{1:t})| (\vx_{1:t}, \theta) \sim \gauss(\vm, \vK_t^{\theta}),
\end{equation*}
where $\vm$ denotes the vector of prior mean evaluated at the data, \emph{i.e.}\
$\left[\vm\right]_i=m(\vx_i)$;
and $\left[\vK_t^{\theta}\right]_{ij}=k^{\theta}(\vx_i,\vx_j)$ is the
covariance matrix between observed data points, parameterised by $\theta$.
For convenience, in this work we use a constant prior mean function.

The choice of covariance function is important as it governs the smoothness of the
function.
While squared exponential kernels are popular, we opt for the Mat\'ern(5/2) kernel
with automatic relevance determination:
\begin{align}
    k^{\theta}(\vx,\vx')
    &= \theta_0 \exp(-\sqrt5 r) (1+\sqrt{5}r+\tfrac53r^2),
    \label{eqn:matern}
\end{align}
where $r = (\vx-\vx')\T \mathbf{\Lambda}^{-1}(\vx-\vx')$ and 
$\mathbf{\Lambda}$ is the diagonal matrix of squared length scale parameters
$\theta_{1:d}$.
Let $\theta$ denote the $d+1$ hyper-parameters which completely characterize
our kernel.
The Mat\'ern(5/2) makes less stringent smoothness assumptions than the squared
exponential kernel and is thus a better fit for heteroscedastic Bayesian
optimisation.

Given the noise-corrupted observations $\data_t$,
the joint distribution of these observations and an arbitrary point $\vx$ is:
\begin{equation*}
    \left[\begin{matrix}
        \vy\\
        f(\vx)
    \end{matrix}\right] \left| \theta\sim
    \gauss\left(
    \left[\begin{matrix} \vm \\ 
    	  m(\vx)\end{matrix}\right],
    \left[\begin{matrix}
        \vK^{\theta}_t+\sigma^2\vI & \vk^{\theta}_t(\vx) \\
        \vk^{\theta}_t(\vx)\T & k^{\theta}(\vx, \vx)
    \end{matrix}\right]
    \right), \right. 
\end{equation*}
where $\left[\vy\right]_i = y_i$ and
$\left[\vk^{\theta}_t(\vx)\right]_i=\vk^{\theta}(\vx_i,\vx)$.
By conditioning on the observed $\vy$, the posterior predictive distribution of
an arbitrary point $\vx$ is marginally Gaussian with mean and variance
\begin{align}
    \mu_t(\vx; \theta)
	&= m(\vx)
	+ \vk^{\theta}_t(\vx)\T(\vK^{\theta}_t+\sigma^2\vI)^{-1}(\vy-\mathbf{m}),
    \label{eqn:mean}
    \\
    \sigma_t^2(\vx; \theta)
    &= k^{\theta}(\vx,\vx)
    - \vk^{\theta}_t(\vx)\T (\vK^{\theta}_t+\sigma^2\vI)^{-1} \vk^{\theta}_t(\vx),
    \label{eqn:std}
\end{align}
respectively.

\subsection{Acquisition functions}

Having specified a distribution to capture our beliefs about the behaviour of the function, as well as a mechanism to update it at each step, we define an acquisition function $\alpha(\cdot|\data_t)$ for choosing the next evaluation point 
\begin{equation*}
\vx_{t+1} = \argmax_{\vx\in\calX}\alpha(\vx|\data_t).
\end{equation*}
The acquisition function must trade-off exploration and exploitation to ensure that the location of the global maximum (or minimum) is found in as few steps as possible.

Although many acquisition strategies have been proposed (see for example~\cite{Mockus:1982,Jones:2001,Hoffman:2011,Hennig:2012,Snoek:2012,Hoffman:2014,Wang:2014aistats,Shahriari:2014}), the expected improvement (EI) criterion remains a default choice in popular Bayesian optimisation software packages, such as SMAC and Spearmint~\citep{Hutter:smac,Snoek:2012}. If
we let \[\vx^+_t=\argmax_{i\leq
t}f(\vx_i; \theta)\] denote the current \emph{incumbent}, the EI acquisition function can be
written in closed form as
\begin{align*}
    \alpha^\textrm{EI}_{\theta}(\vx|\data_t)
    &= \E[\max\{0,f(\vx) - f(\vx^+)\}|\data_t]\\
    &= \sigma_t(\vx; \theta)[z\Phi(z) + \phi(z)]
    \label{eqn:eideterministic}
\end{align*}
with
\begin{equation*}
	z = \frac{\mu_t(\vx; \theta) - f(\vx^+)}{\sigma_t(\vx; \theta)},
\end{equation*}
and $\phi$, $\Phi$ representing the standard
normal density and distribution functions respectively. In the special case of $\sigma_t(\vx; \vtheta) =0$, we set $\alpha^\textrm{EI}_{\vtheta}(\vx|\data_t)= 0$. The expected improvement is best understood as a family of one-step-decision heuristics~\cite{Brochu:2009}, with many members in this family.
\begin{figure*}[!t]
    \centering
    \begin{subfigure}{.3\linewidth}
        \centering
        \includegraphics[width=\textwidth]{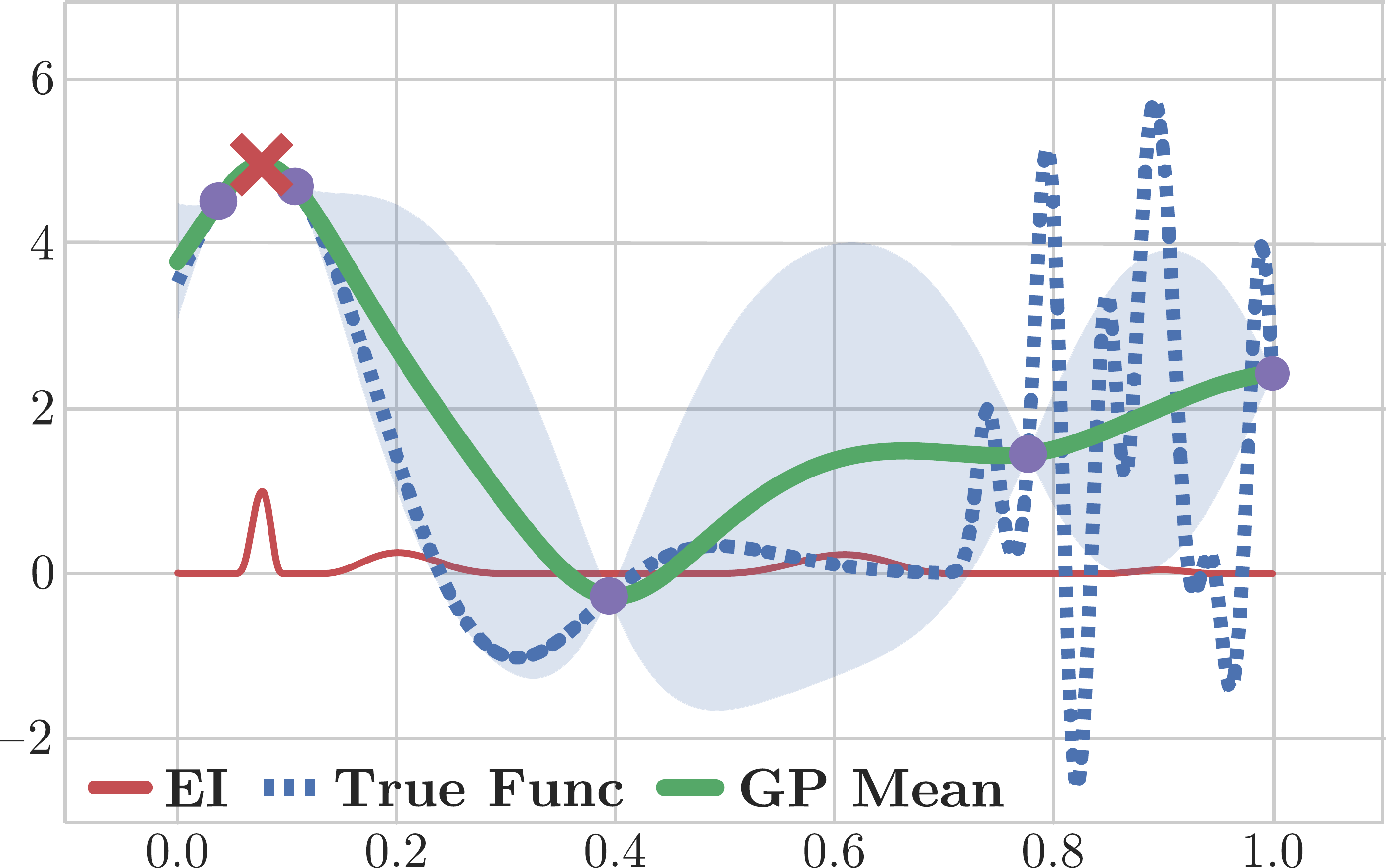}
       \caption{BO Iteration 8}
    \end{subfigure}
    \begin{subfigure}{0.04\linewidth}
        \centering
        \includegraphics[width=0.3\textwidth]{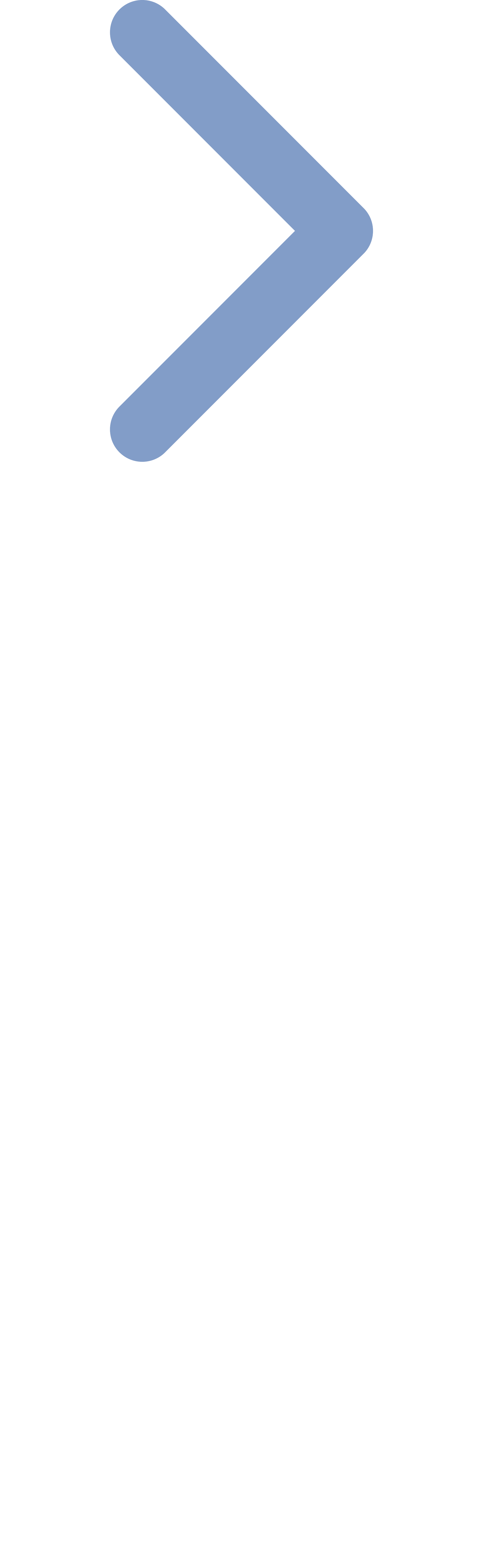}
    \end{subfigure}
    \begin{subfigure}{.3\linewidth}
        \centering
        \includegraphics[width=\textwidth]{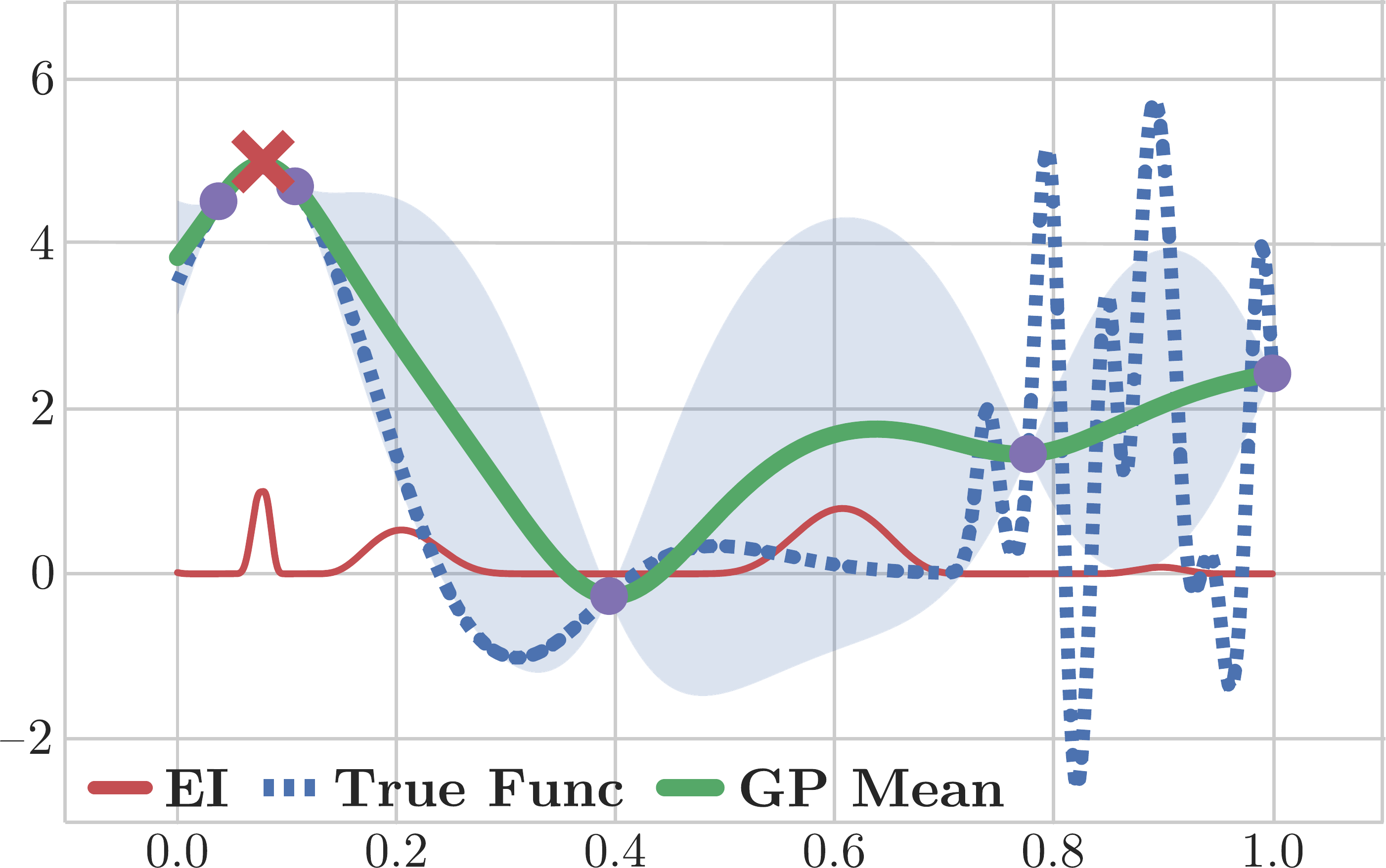}
       \caption{BO Iteration 15}
    \end{subfigure}
    \begin{subfigure}{0.04\linewidth}
        \centering
        \includegraphics[width=0.3\textwidth]{artwork/arrow_right}
    \end{subfigure}
    \begin{subfigure}{.3\linewidth}
        \centering
        \includegraphics[width=\textwidth]{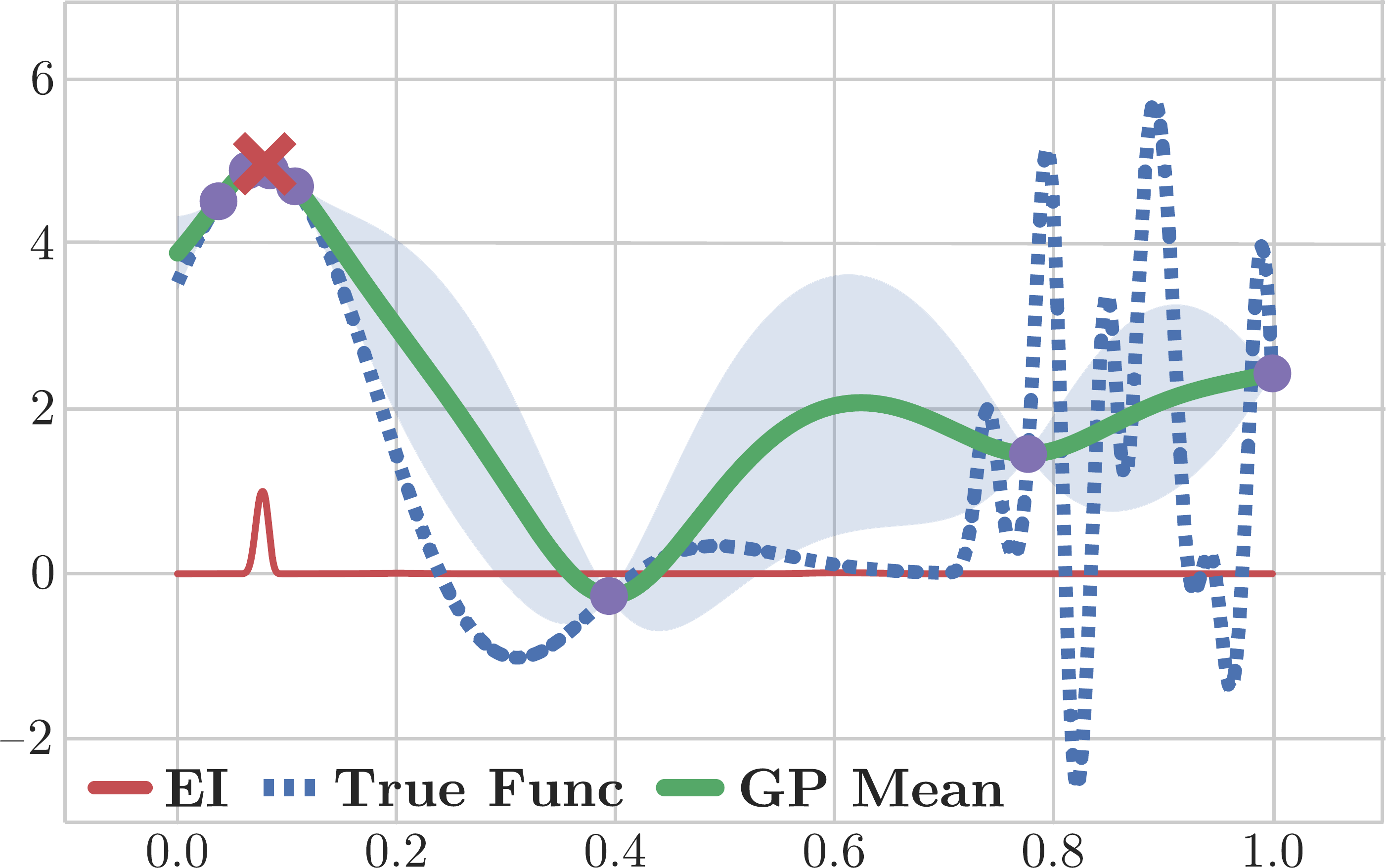}
       \caption{BO Iteration 35}
    \end{subfigure}\\
    
    \begin{subfigure}{.3\linewidth}
        \centering
        \includegraphics[width=\textwidth]{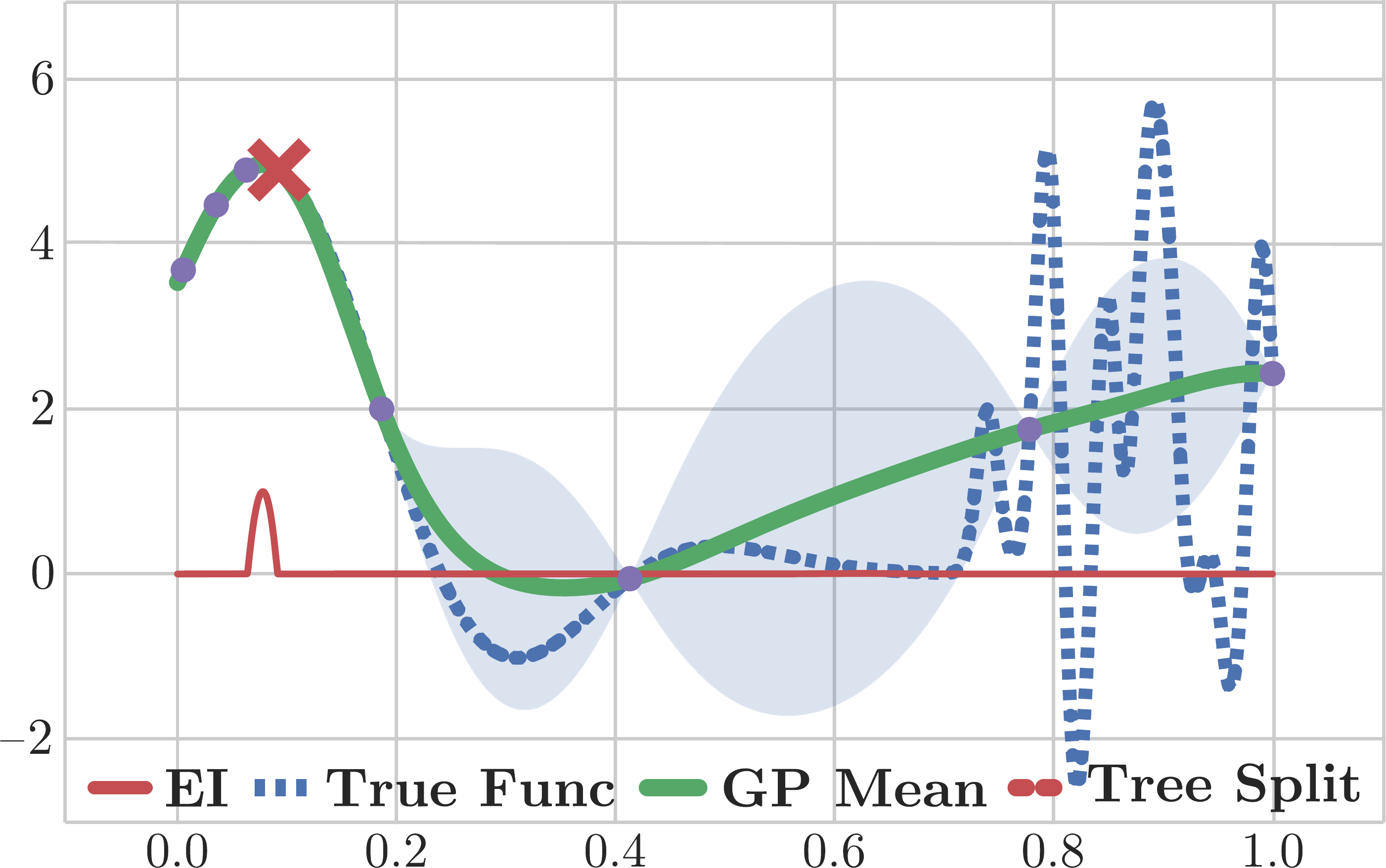}
        \caption{HTBO Iteration 8}
    \end{subfigure}
    \begin{subfigure}{0.04\linewidth}
        \centering
        \includegraphics[width=0.3\textwidth]{artwork/arrow_right}
    \end{subfigure}
    \begin{subfigure}{.3\linewidth}
        \centering
        \includegraphics[width=\textwidth]{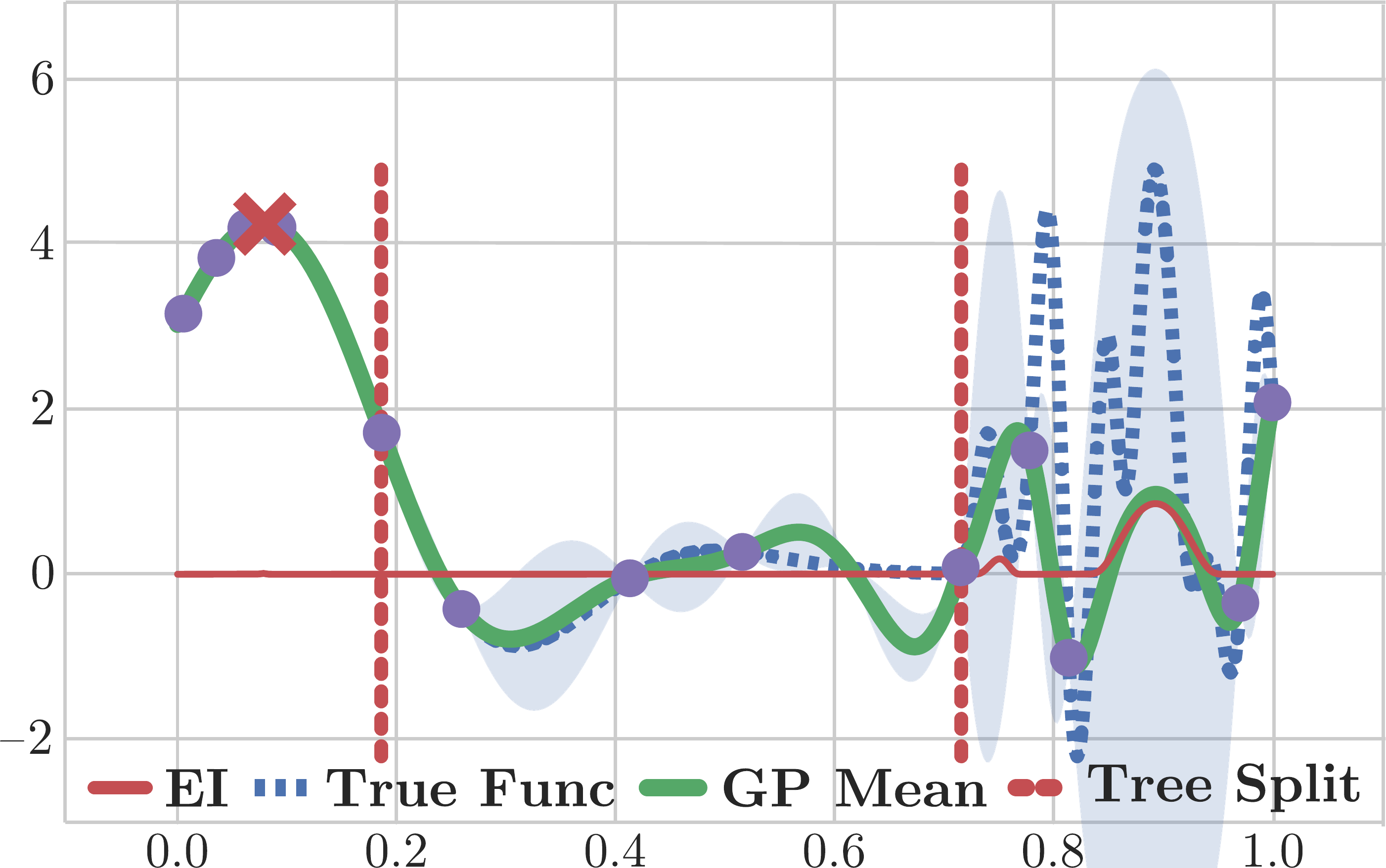}
        \caption{HTBO Iteration 15}
    \end{subfigure}
    \begin{subfigure}{0.04\linewidth}
        \centering
        \includegraphics[width=0.3\textwidth]{artwork/arrow_right}
    \end{subfigure}
    \begin{subfigure}{.3\linewidth}
        \centering
        \includegraphics[width=\textwidth]{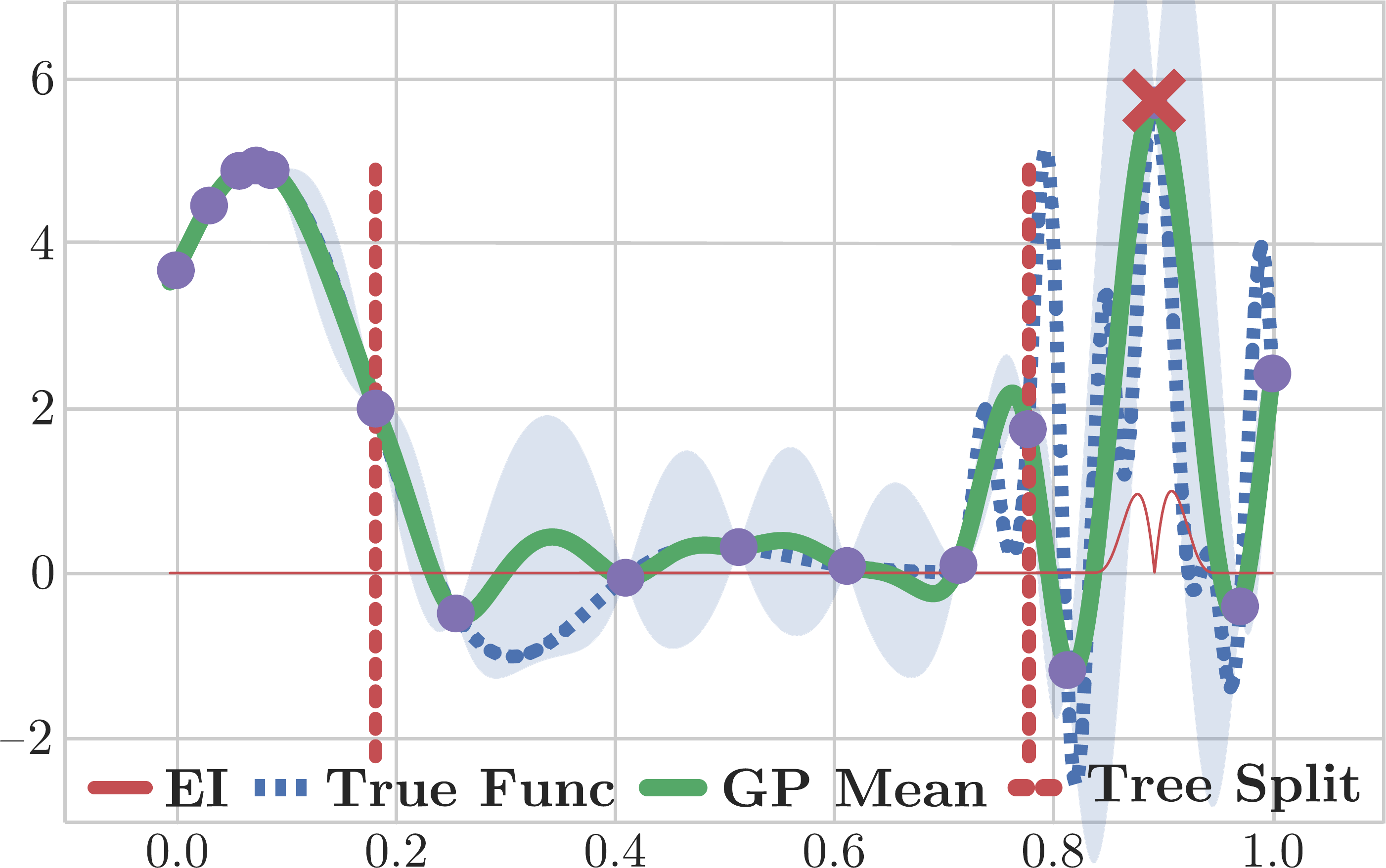}
        \caption{HTBO Iteration 16}
    \end{subfigure}
	\caption{Comparison between standard BO (a-c) and the proposed HTBO method (d-f). HTBO is able to find the maximum in 16 iterations, while BO is not able to overcome heteroscedasticity and over-samples one of the local maxima.}
	\label{fig:tree_split} 
\end{figure*}

\subsection{Related work}

Several approaches have been proposed to manage heteroscedasticity with Gaussian processes. \cite{sampson1992nonparametric} attempted to project inputs into a latent space that is stationary. This approach was later extended by~\cite{schmidt2003bayesian}. A latent space representation in higher dimensions was also proposed by~\cite{bornn2012modeling}. Others such as~\cite{higdon1999non} and~\cite{williams2006gaussian} have tried to model heteroscedasticity directly with the choice of covariance function.
In 2005,~\citeauthor{gramacy2005bayesian} proposed a treed GP model to attack non-stationarity.
While this work, as well as, the work of~\cite{Dunson:2012}, are the closest to ours, both were developed for modelling functions and not for global optimisation under a limited number of observations.

Warping is another popular approach for dealing with non-stationarity~\citep{snelson2004warped,adams2008gaussian}. Recently,~\cite{snoek:2013b} proposed an input warping technique, using a parameterised Beta cumulative distribution function (CDF) as the warping function. The goal of input warping is to transform non-stationary functions to stationary ones by applying a Beta CDF mapping $w_d(\cdot)$ to each dimension $d$. The new covariance becomes $\kappa(w(\vx), w(\vx'))$. We have found that input warping can lead to remarkable improvements in automatic algorithm configuration. However, the Beta CDF transformation has limitations, which we address in this paper by using a treed approach.

\begin{figure*}[t!]
    \centering 
	\begin{minipage}[b]{.35\linewidth} \centering 
		\includegraphics[width=1.\textwidth]{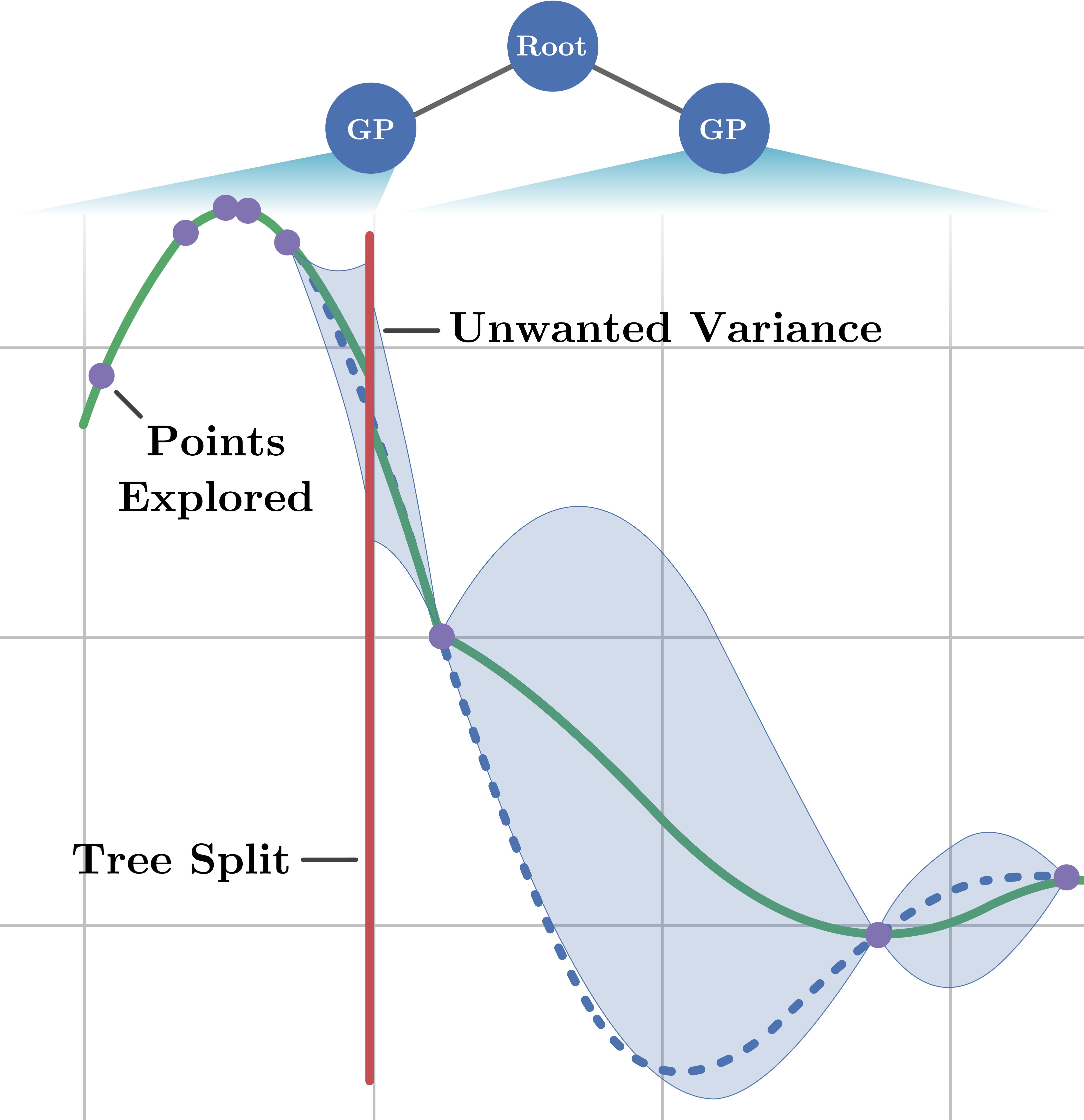} \subcaption{CART Splitting}
	\end{minipage}
	\begin{minipage}[b]{.01\linewidth}
	\end{minipage}
	\begin{minipage}[b]{.35\linewidth} \centering 
		\includegraphics[width=1.\textwidth]{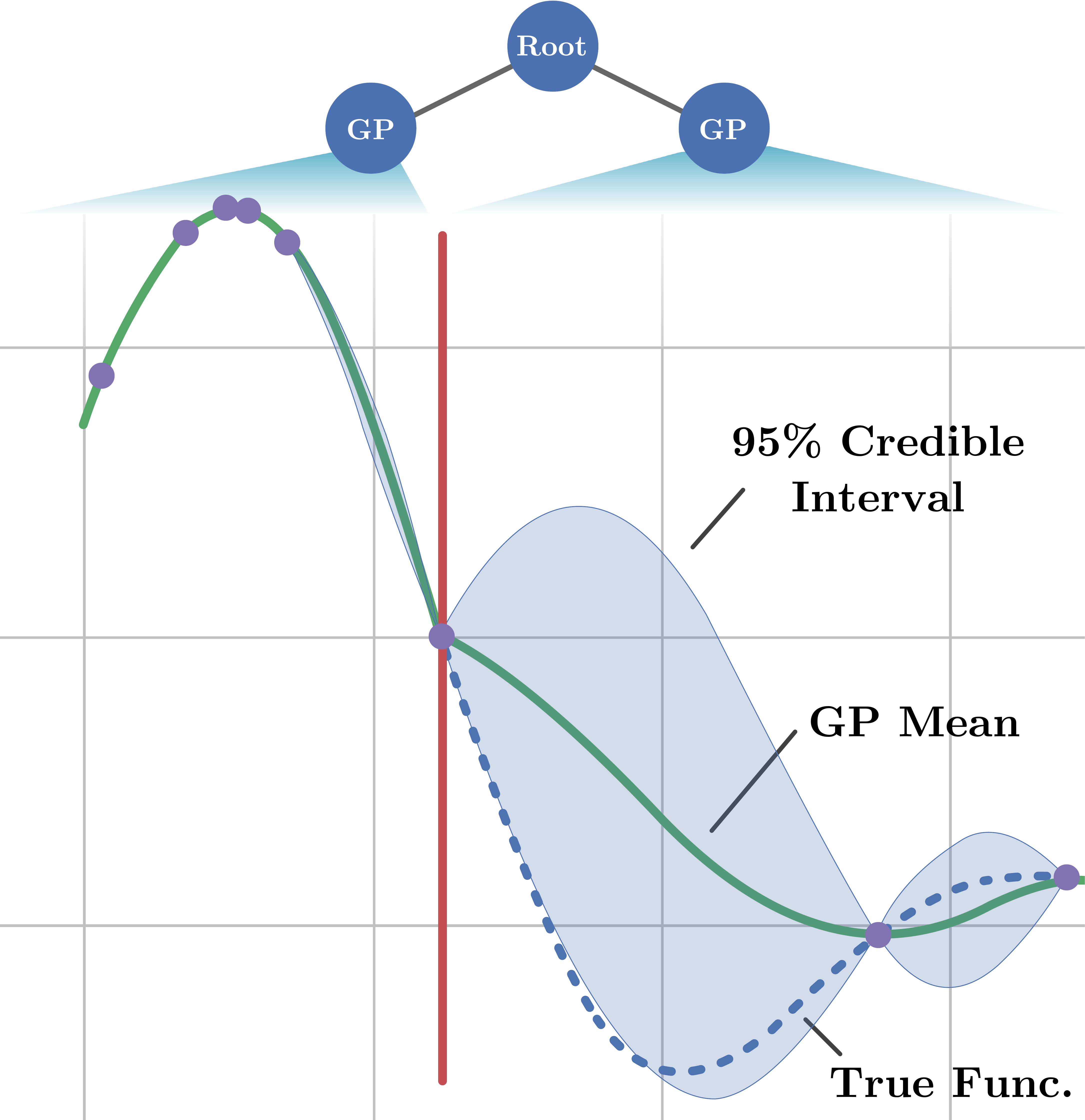} \subcaption{Proposed approach}
	\end{minipage}
	\caption{Comparison between conventional CART splitting and our proposed splitting method.
	The proposed splitting method reduces the variance on the boundaries thus reducing waste of samples.}
	\label{fig:tree_unwanted_variance}
\end{figure*}
	
\section{Treed Bayesian Optimisation}

\subsection{Constructing the tree structure}
	
Our proposed heteroscedastic treed Bayesian optimisation (HTBO) method is based on classification
and regression trees (CART), a decision tree model of~\cite{breiman1984classification}.
A Decision tree may be understood in terms of a sequence of binary tests applied to an input $\vx$, which determines the path followed by $\vx$ from the root of the tree to a leaf. Each
node has a function of the form $h(\vx) > \tau$, where $h$ extracts a coordinate (feature) of $\vx$ and compares it to a threshold $\tau$. The tree is constructed in a recursive manner by choosing splits on features and thresholds so as to reduce uncertainty~\citep{matheson2014empirical}.

We can measure uncertainty in a node $A$ using the empirical mean squared error: \[\text{U}(A) = \frac{1}{|A|} \sum_{y_i \in A}(\bar{y}_A - y_i)^2,\] where $\bar{y}_A$ is the average of the output values in $A$. We could also use the entropy of the GPs in each node, but we found this alternative uncertainty measure to require much more computation without leading to better performance.

The optimal splits on features and thresholds are the ones that reduce uncertainty the most when splitting node 
$A$ into $A'_{h,\tau}$ and $A''_{h,\tau}$. They are obtained by optimising the following reduction in uncertainty objective:
\begin{align}
\text{I}(A, A'_{h,\tau} , A''_{h,\tau} ) =  \text{U}(A) & - \frac{|A'_{h,\tau}|}{|A|} \text{U}(A'_{h,\tau} ) \nonumber \\ &- \frac{|A''_{h,\tau}|}{|A|} \text{U}(A''_{h,\tau} ).
\end{align}
In CART, the splitting threshold $\tau$ of feature $h(\vx)$ is the midpoint of two points $(\vx_i,\vx_j)$, which is convenient for constant predictions as $\vx_i$ will go to the left child and $\vx_j$ to the right one respectively. However, in the proposed approach this would create unwanted variance in the gap between $\vx_i$ and $\vx_j$, as one GP will have to cover the unknown space from $\vx_i$ to $\tau$ and another GP the unknown space from $\tau$ to $\vx_j$. This antagonises the goal of minimising the conditional variance in Bayesian Optimisation~\citep{Brochu:2007} as shown in the left plot of Figure~\ref{fig:tree_unwanted_variance}.

We solve the problem of the unwanted variance, by placing $\tau$ exactly at one of the points $\vx_i$, and let $\vx_i$ belong to both children nodes, as shown on the right hand side of Figure~\ref{fig:tree_unwanted_variance}. This splitting strategy is essential for Bayesian optimisation to work well with treed GPs.

\newpage
\subsection{Learning hyper-parameters}

Whereas other treed GP models propose partitioning the data for computational
efficiency~\citep{Gramacy:2004,Bui:2014} or capturing multiple scales in the
data~\citep{Dunson:2012},
our approach is designed to learn non-stationarity in the random process that
generated the objective function.
Therefore, it is paramount that our model is capable of learning different
hyper-parameters for each leaf. 

Maximum likelihood is a common technique for obtaining a point estimate of the
hyper-parameters. In GP regression, the log-marginal-likelihood can be expressed
analytically as follows
\begin{align}
    2\log p(\vy|\vx_{1:t},\theta) =& -\vy\T (\vK^{\theta}_{t} + \sigma^2 I)^{-1} \vy
    \nonumber
    \\
    &-\log |\vK^{\theta}_{t} + \sigma^2 I| - t \log(2\pi).
\end{align}
A straightforward implementation of this approach for our purpose would independently maximise the log-marginal-likelihood of the data in each leaf.
When leaves have very few data points, however, maximising the likelihood  could severely underestimate the length scale hyper-parameters.

\begin{figure}[h]
  \centering
  \vspace{0pt}
  \includegraphics[width=1.0\linewidth]{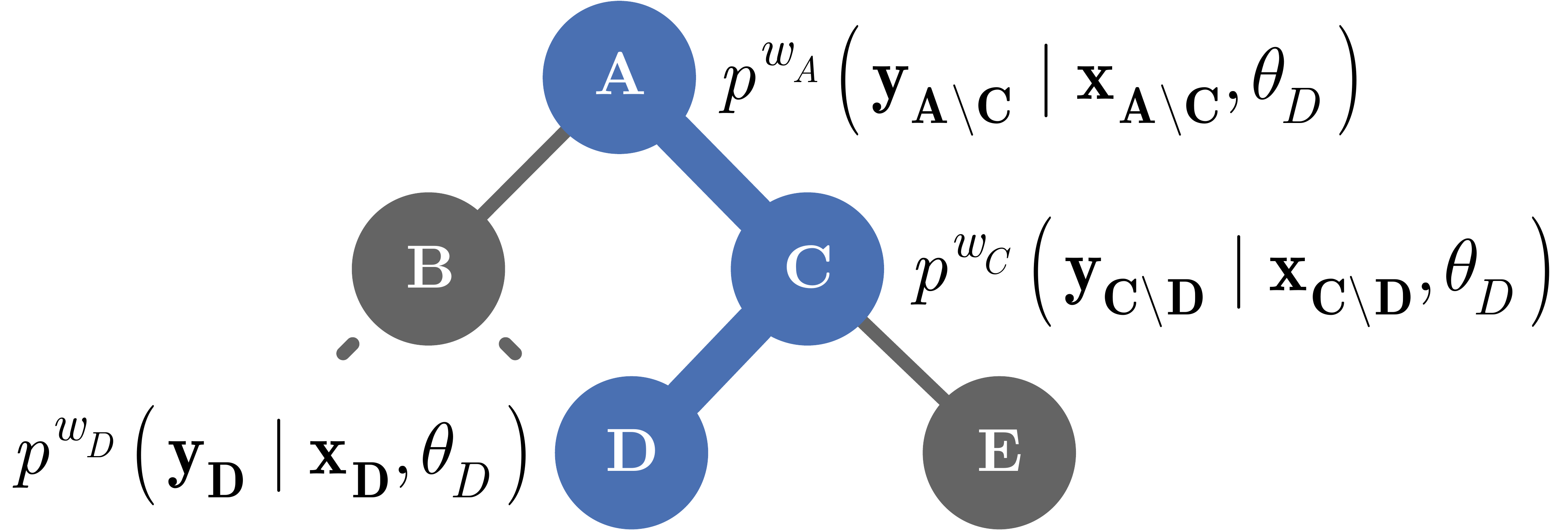}
  \caption{Tree-structured hyper-parameter estimation for the GP at leaf
  	node $D$. For simplicity, the nodes in this example are indexed by
  	letters rather than integers (as in the text).}
	\label{fig:tree_prior}
\end{figure}

In this section, we propose a way of aggregating information from different levels of the tree hierarchy. Given a tree-structured partition of current observations $\data_t$ constructed as per
the previous section, let nodes be integer indexed, starting with the root node $0$.
Let $\vy_{(i)}$ denote the data in node $i$ and $\vy_{(i\setminus j)}$ denote the data
in node $i$ excluding the data in node $j$, and similarly for $\vx_{(i)}$ and
$\vx_{(i\setminus j)}$.
Furthermore, let $\delta^i$ return the depth of node $i$ such that $\delta^0=0$ and
let $\rho^i$ return the ordered list of nodes in the path from the node $i$ to the
root $0$, such that $\rho^i_0 = i$ and $\rho^i_{\delta^i} = 0$, and finally let
$\calL_t$ denote the set of leaves of the tree.
\newpage
Suppose we are interested in estimating the hyper-parameters of the GP associated
with leaf $j\in\calL_t$, consider the following marginal pseudo-likelihood
decomposition
\begin{multline}
	p(\vy | \vx_{1:t}, \theta)
	= p\Big(\vy_{(j)} \bigcup_{i=1}^{|\rho_j|} \vy_{(\rho^j_i\setminus\rho^j_{i-1})}
		\Big| \vx_{1:t}, \theta\Big)
	\\
	\approx
	p(\vy_{(j)} | \vx_{(j)}, \theta) \prod_{i=1}^{|\rho^j|}
	p\Big(\vy_{(\rho^j_i\setminus\rho^j_{i-1})}
		\Big| \vx_{(\rho^j_i\setminus\rho^j_{i-1})},\theta\Big).
\end{multline}
This decomposition implicitly assumes certain conditional independencies between
nodes in the tree, given $\theta$.
Each factor corresponds to a node along the path $\rho_j$ of the leaf $j$ in question.
However, the factors are all equally weighted, so that the estimated hyper-parameter
$\hat\theta_{\mathrm{ML}}$ fits all of the data equally.
Consider instead, for each leaf $j\in\calL_t$, a \emph{weighted} marginal
pseudo-likelihood decomposition
\begin{multline}
	p(\vy | \vx_{1:t}, \theta)
	\approx
	p^{w^j_0}(\vy_{(j)} | \vx_{(j)}, \theta)
	\\
	\times \prod_{i=1}^{|\rho^j|}
	p^{w^j_i}\Big(\vy_{(\rho^j_i\setminus\rho^j_{i-1})}
		\Big| \vx_{(\rho^j_i\setminus\rho^j_{i-1})},\theta\Big),
\end{multline}
where the importance of each factor depends on its corresponding node's depth relative to leaf $j$. A simple example of this factor decomposition is depicted in Figure~\ref{fig:tree_prior}.

The weighted likelihood was inspired by the work of~\citep{newton1994approximate} and paves us the way to produce samples from the posterior of interest.
Finally, the weights $w^j$ are selected so that the factor corresponding to the leaf~$j$ is squared and weights decrease harmonically along the path~$\rho^j$:
\begin{equation}
	w^j_i = \frac2{1 + \delta^j - \delta^i}.
\end{equation}

So far, for ease of presentation, we have focused our attention on maximising the
marginal likelihood (also known as empirical Bayes or maximum likelihood II).
However, it is straightforward to adopt a more Bayesian approach by
prescribing a prior $p(\theta)$ and inferring the hyper-parameters of
leaf $j$ by targeting the following unnormalised posterior
\begin{multline}
	p(\theta | \vx_{1:t}, \vy)
	\propto
	p(\theta)p^{w^j_0}(\vy_{(j)} | \vx_{(j)}, \theta)
	\\
	\times \prod_{i=1}^{|\rho^j|}
	p^{w^j_i}\Big(\vy_{(\rho^j_i\setminus\rho^j_{i-1})}
		\Big| \vx_{(\rho^j_i\setminus\rho^j_{i-1})},\theta\Big),
\end{multline}
using Markov chain Monte Carlo (MCMC).

\subsection{Heteroscedastic treed Bayesian optimisation}
At each iteration of Bayesian optimisation, the decision tree is reconstructed. In doing so, we ensure that there is a minimum number of data points per leaf (5 in our experiments). Subsequently, we estimate the hyper-parameters as discussed in the previous section. The same approach is also used in the estimation of the mean $\mu_t$ and the kernel amplitude $\theta_0$ of the GP. Once the GPs have been fit to the data in each leaf, we use their statistics to construct the EI acquisition function. This function is then optimised using a $d$-dimensional Sobol Grid with $20,000$ points as in the package Spearmint.

It is interesting to note that, \textsc{HTBO} does not place explicit requirements on the GP leaves, so long as, we can compute the log-marginal-likelihood and EI efficiently.
As a result, not only could we use standard GPs but also
GPs with warping~\citep{snoek:2013b} as leaves.
When employing warping, the $\alpha$ and $\beta$ parameters of the Beta CDF are estimated using the proposed hierarchical prior.
We refer to \textsc{HTBO} with Warping as \textsc{HTBO~Warp}
and evaluate it in our experiments.

Example iterations of the proposed treed approach against standard Bayesian optimisation on a one-dimensional heteroscedastic function are illustrated in Figure~\ref{fig:tree_split}. While the standard approach fails, the proposed method, hierarchical treed Bayesian optimisation (\textsc{HTBO}), is able to overcome non-stationarity to find the maximum of the objective function.

\section{Experiments}
\label{sec:empirical_results}
In this section we evaluate the proposed methods \textsc{HTBO} 
and \textsc{HTBO~Warp}. 
Comparisons are made against standard 
Bayesian optimisation (\textsc{BO}), as well as, the Bayesian optimisation approach 
with input warping (\textsc{BO~Warp}) of~\citep{snoek:2013b}.
For all the experiments presented in this paper, we used the Mat\'ern(5/2) kernel. 
The GP hyper-parameters of all four approaches are obtained using 
slice sampling.
We use three different sets of experiments for evaluation: 
synthetic functions, algorithm configuration benchmarks, and mineral exploration datasets.
The results are summarised and discussed in Section~\ref{sec:eval}.

\subsection{Synthetic problems}
We first introduce two heteroscedastic synthetic functions.
The first synthetic function, which we refer to as RKHS, 
is shown in Figure~\ref{fig:tree_split}. 
The function is constructed as a weighted sum 
of squared exponential kernel functions with 2 different length scales.
The left hand side of the function is smooth, whereas the right hand size 
jagged. For a more detailed description, as well as, the source code
of the function, please refer to~\cite{rkhsfunction}.

The second synthetic function is a two-dimensional exponential
function from~\cite{gramacy2005bayesian}. 
The precise mathematical expression for the function is:

\begin{equation}
f(x_1, x_2)=x_1 \exp(-x_1^2-x_2^2).	
\end{equation}

\begin{figure}[h]
  \centering
  	\includegraphics[width=0.9\linewidth]{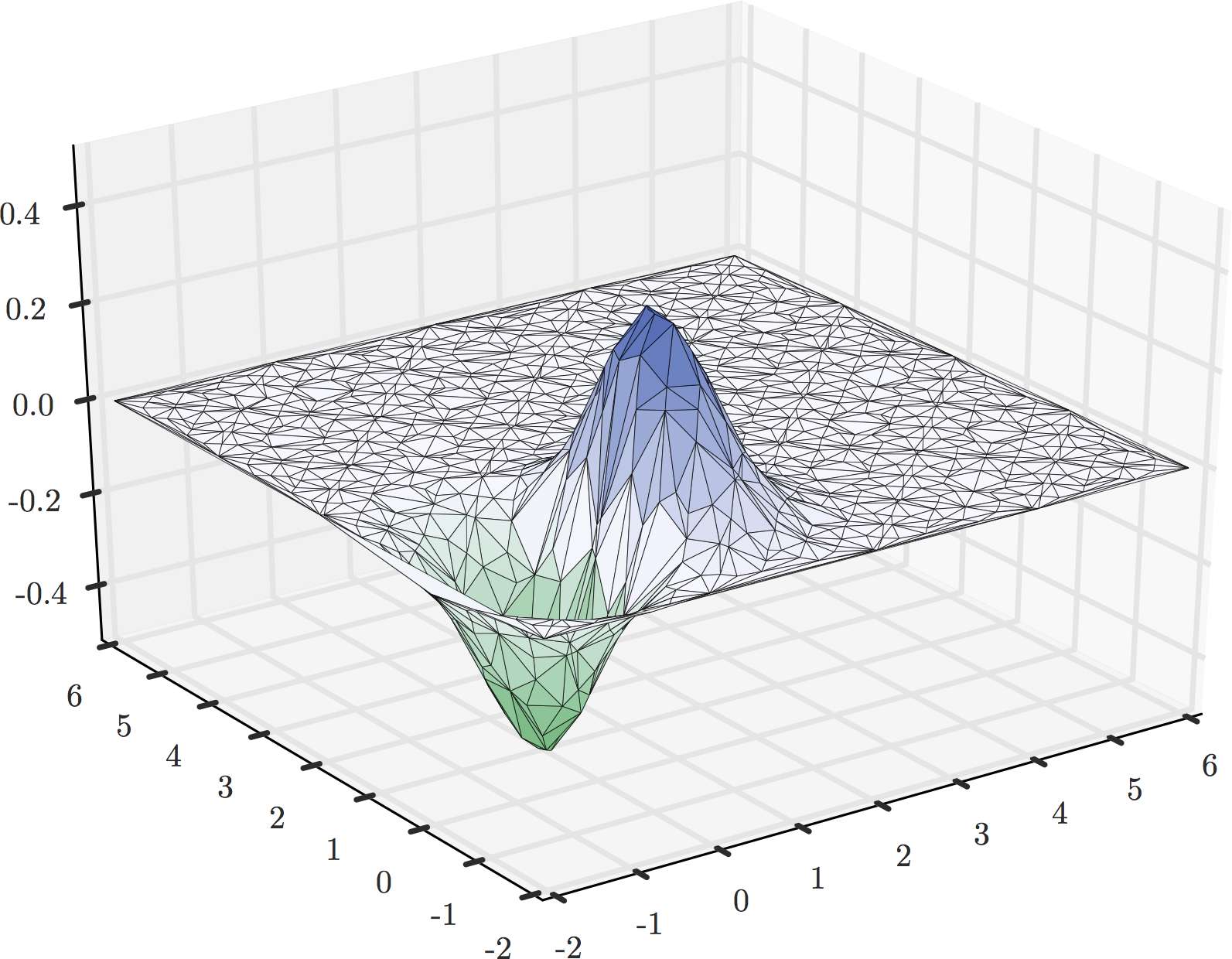}
	\caption{2D Exp function. The function is mostly flat with a peak
	that could be easily missed. }
 	\label{fig:func_gramacy}
\end{figure}

We refer to this function as 2-D Exp. and plot it in Figure~\ref{fig:func_gramacy}.
This function is interesting because it is ``flat'' over most of its domain, with a peak that can be easily missed without careful exploration.

\vspace{0.2in}

\begin{figure*}[!t]
    \centering
    \begin{subfigure}{.45\linewidth}
        \centering
        \includegraphics[width=0.85\textwidth]{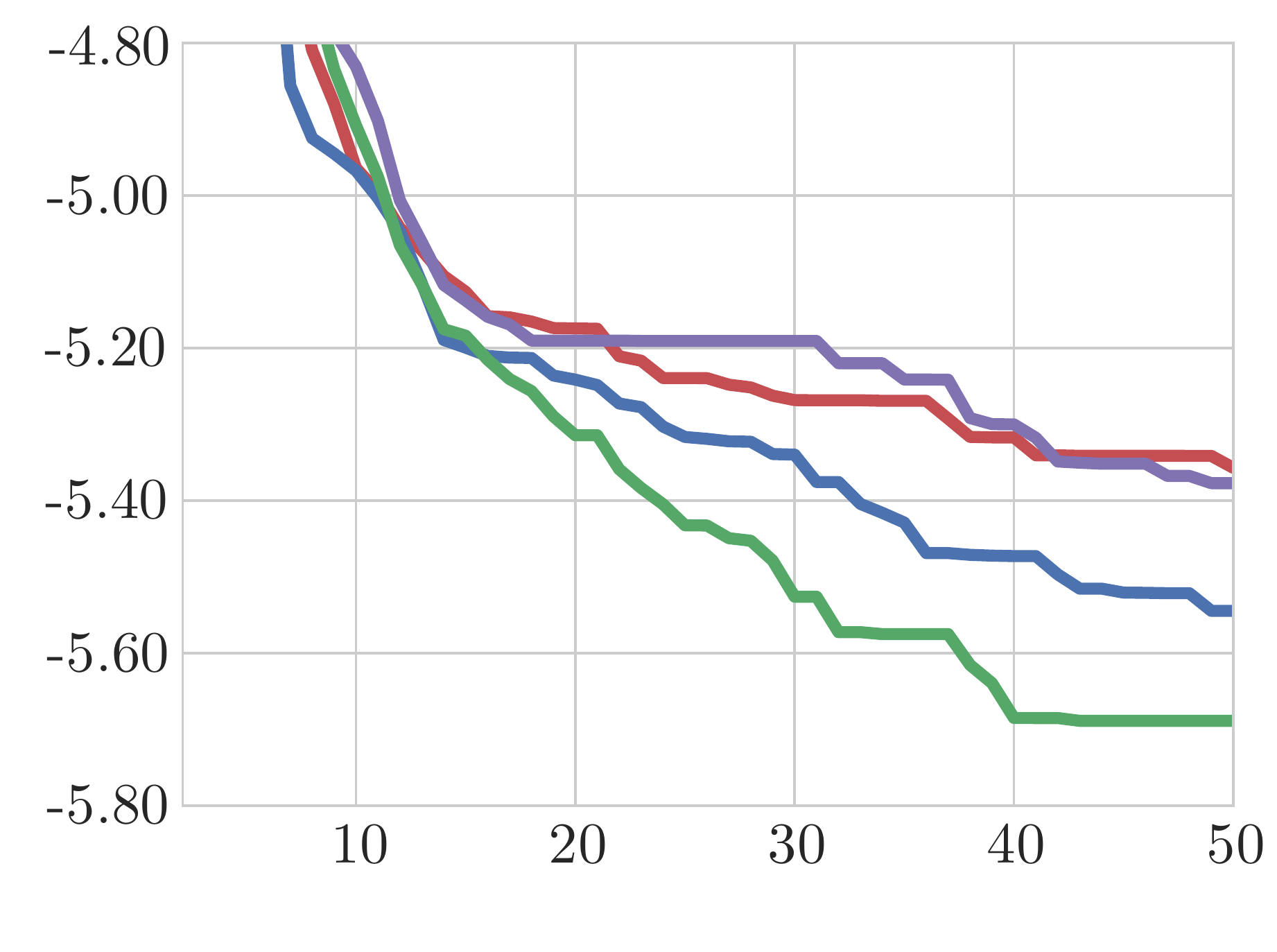}
        \subcaption{RKHS Function}
	\end{subfigure}
    \begin{subfigure}{.45\linewidth}
        \centering
        \includegraphics[width=0.85\textwidth]{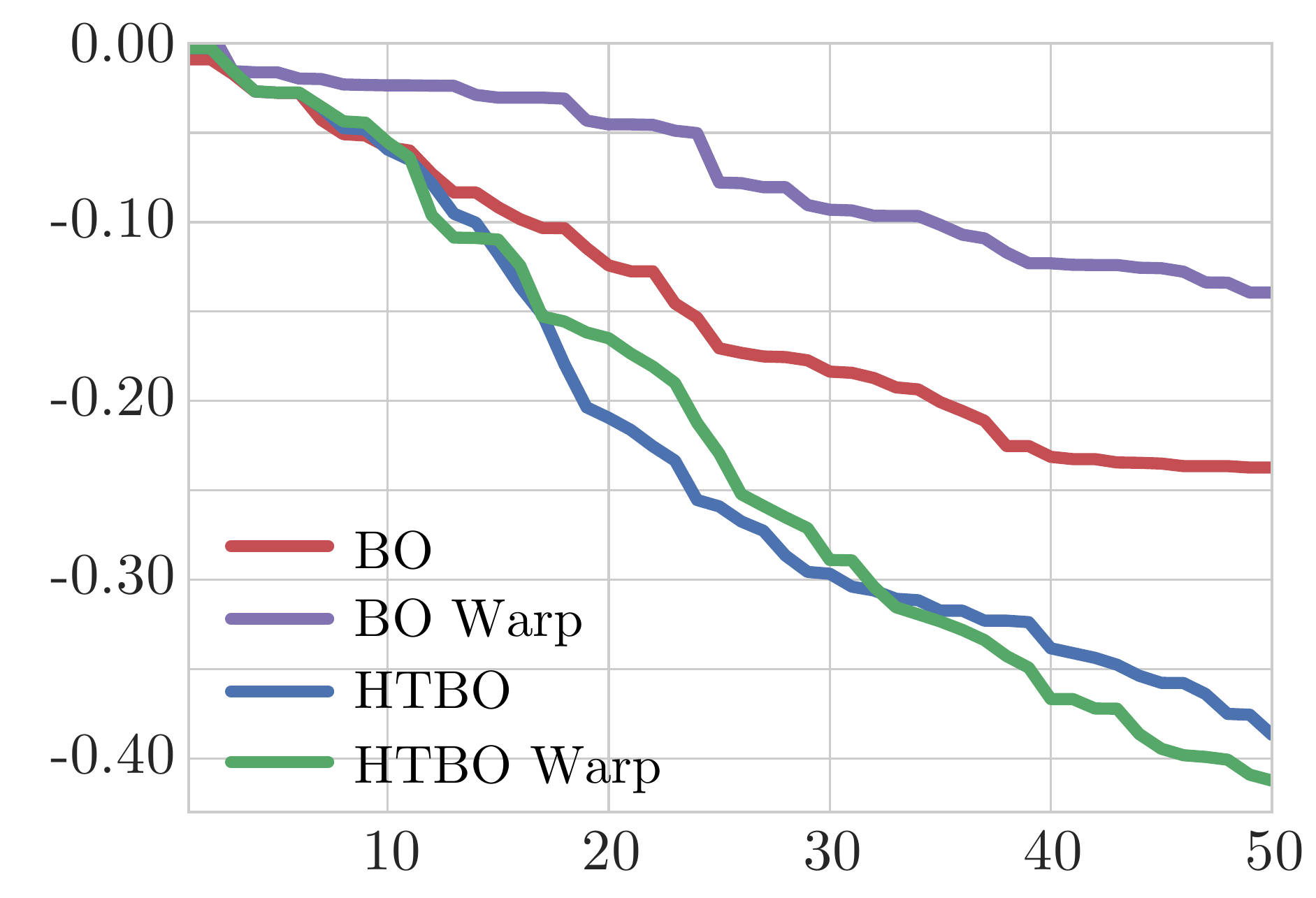}
        \subcaption{2-D Exp. Function}
    \end{subfigure}
    \begin{subfigure}{.45\linewidth}
        \centering
        \includegraphics[width=0.85\textwidth]{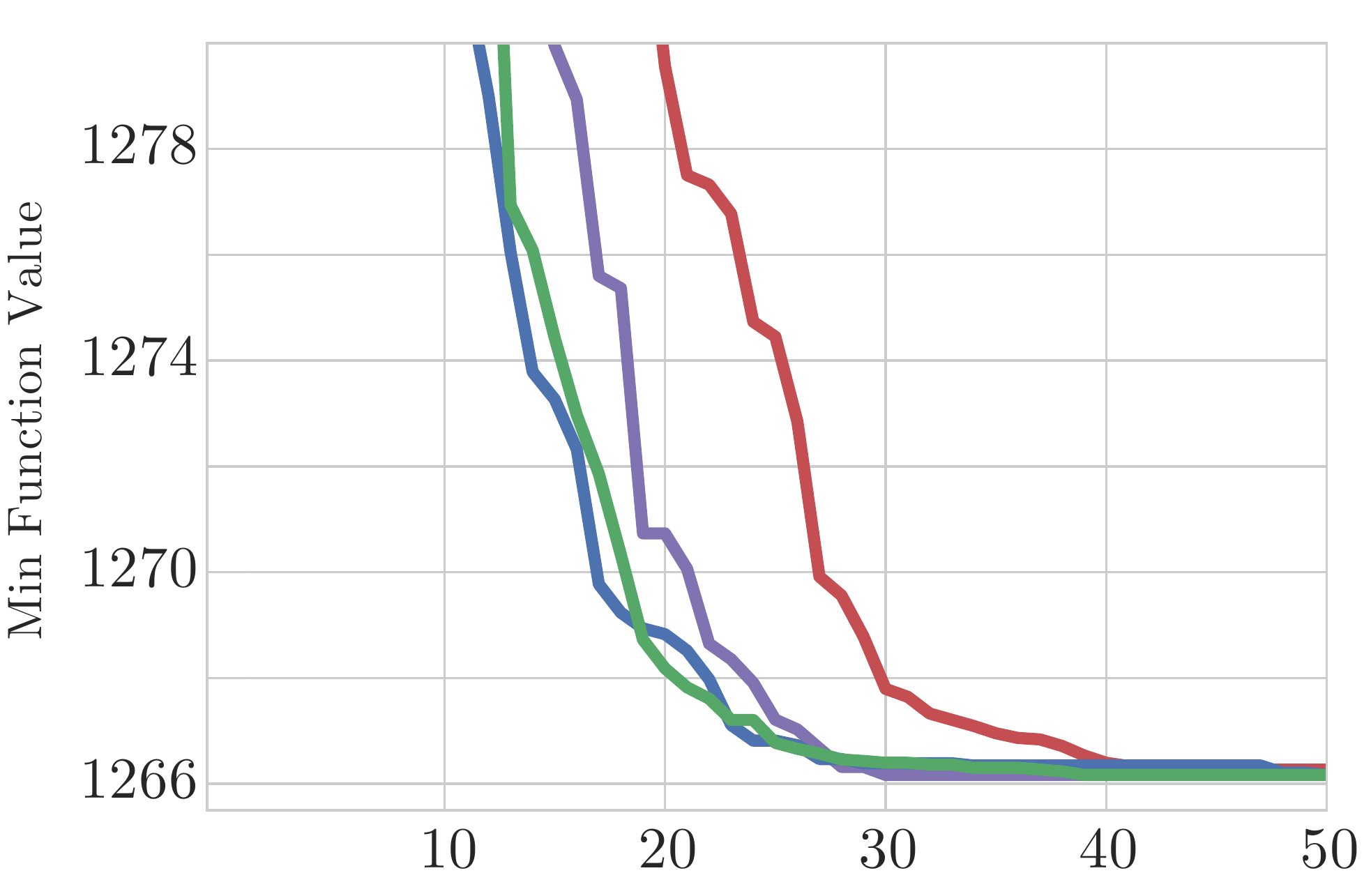}
        \subcaption{Online LDA}
    \end{subfigure}
    \begin{subfigure}{.45\linewidth}
        \centering
        \includegraphics[width=0.85\textwidth]{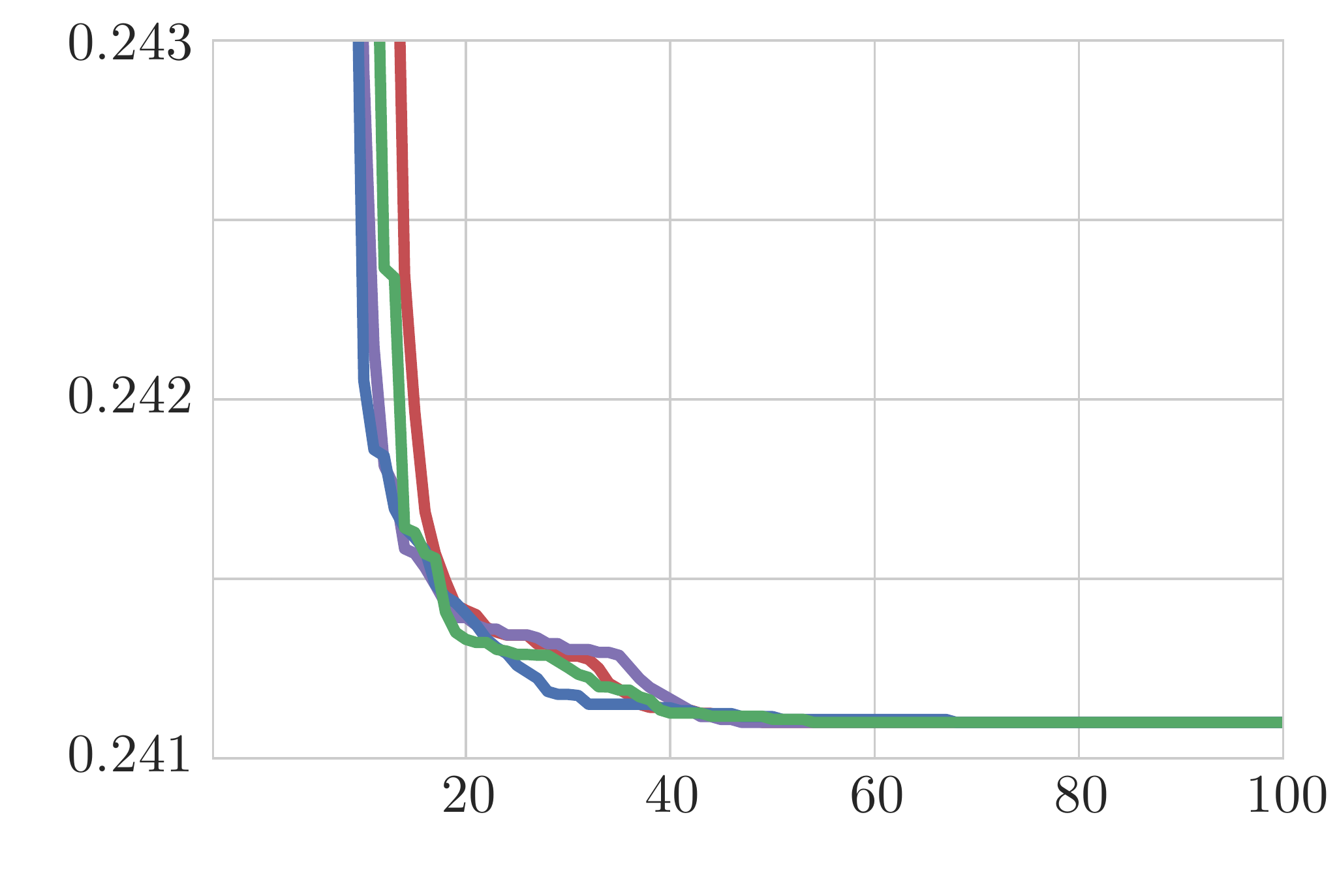}
        \subcaption{Structured SVM}
    \end{subfigure}
    \begin{subfigure}{.45\linewidth}
        \centering
        \includegraphics[width=0.85\textwidth]{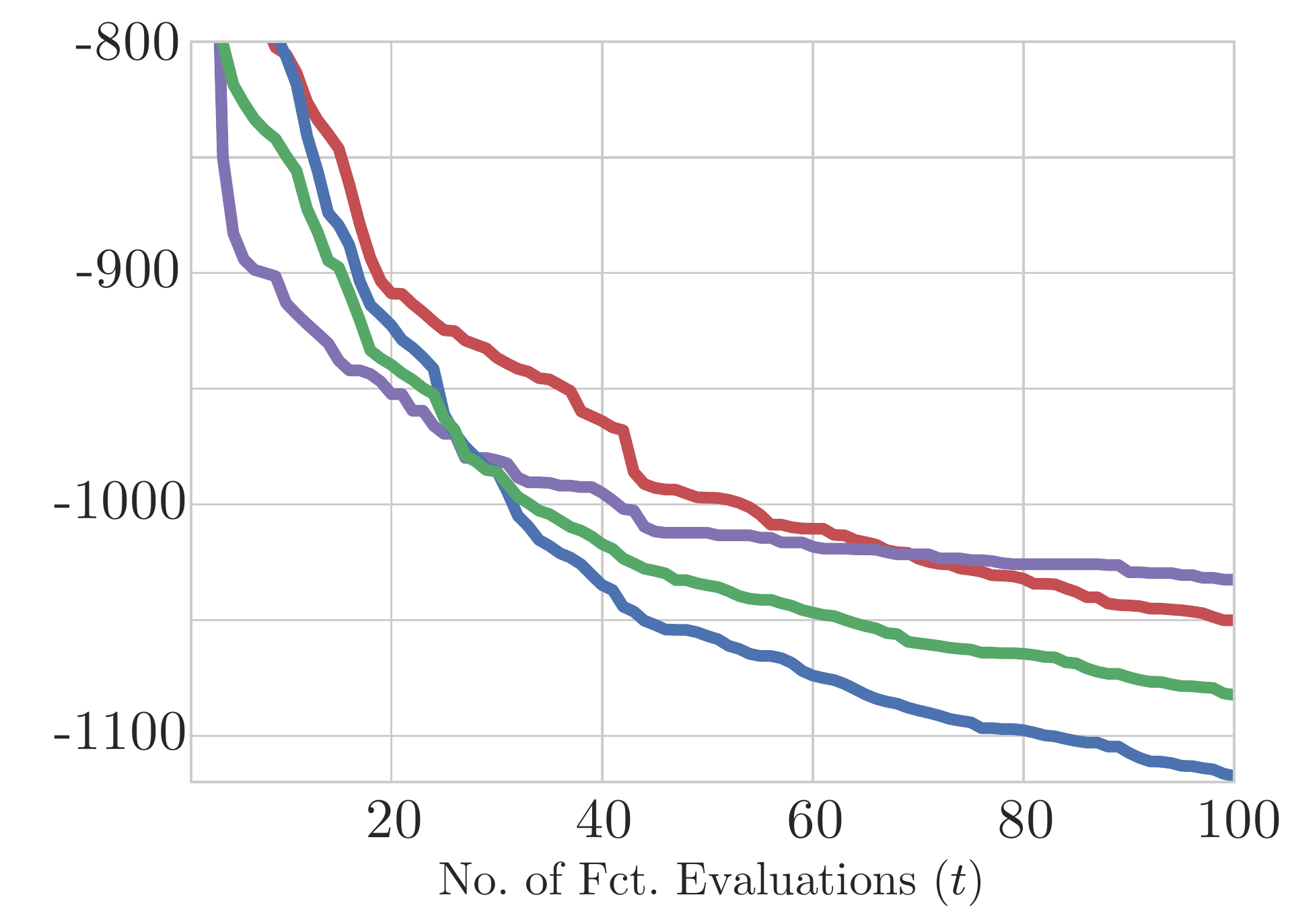}
        \subcaption{Agromet}
    \end{subfigure}
    \begin{subfigure}{.45\linewidth}
        \centering
        \includegraphics[width=0.85\textwidth]{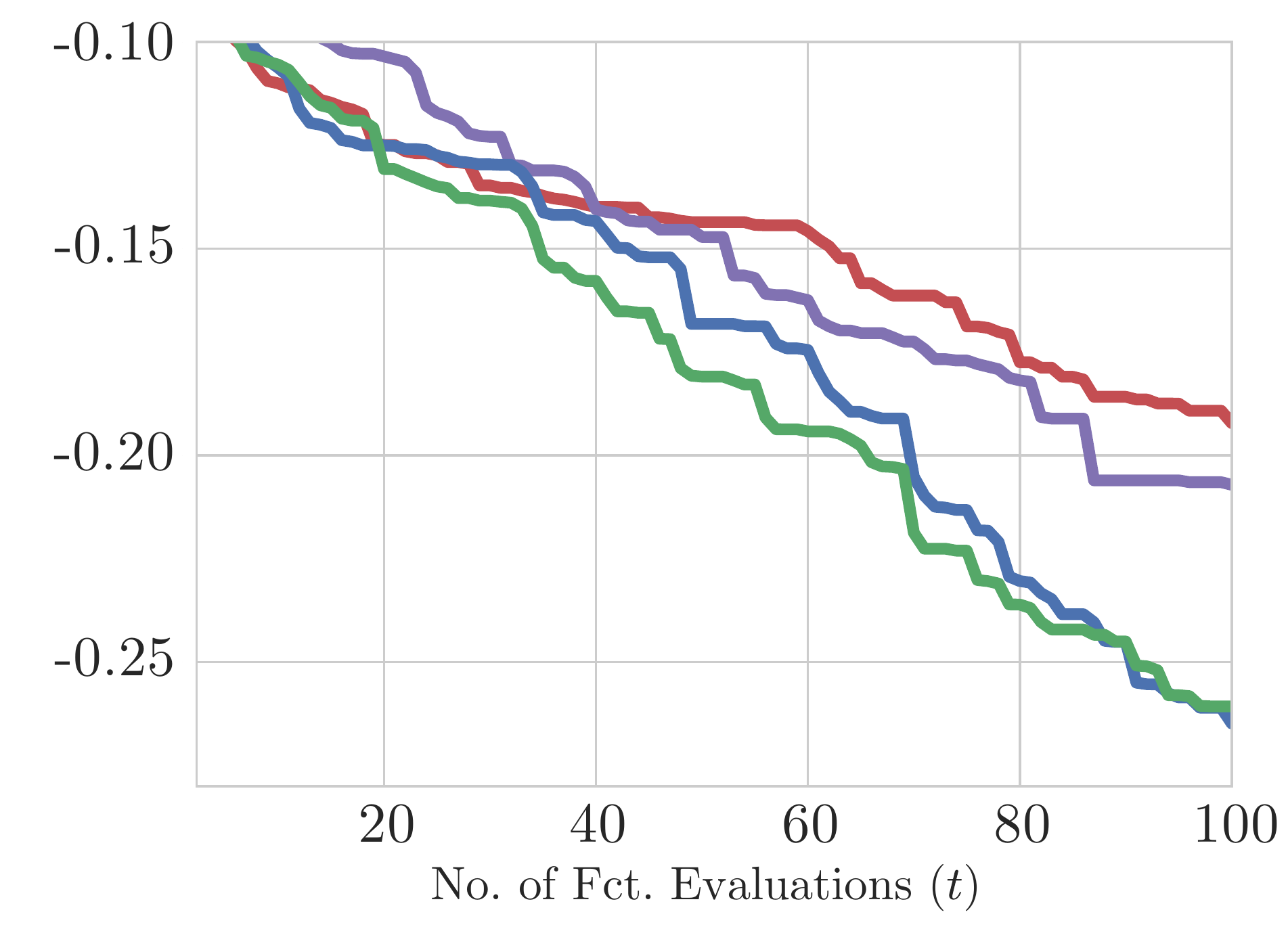}
        \subcaption{Brenda mines}
    \end{subfigure}
	\caption{Performance of BO, BO with input warping (BO~Warp), and the proposed approaches (HTBO and HTBO~Warp) on synthetic functions, algorithm configuration problems and mining problems.}
	\label{fig:results}
\end{figure*}

\newpage
\subsection{Automatic machine learning problems}

Latent Dirichlet Allocation (LDA) is a directed graphical model for documents used in topic modelling tasks, for which~\cite{hoffman2010online} proposed an online learning approach in the variational Bayes paradigm. 
In this experiment we use precomputed performance data for this on-line LDA algorithm on a dataset of 250,000 Wikipedia articles under many parameter settings. Tuning the online LDA algorithm involves choosing the two learning parameters, $\tau_0$ and $\kappa$
as well as a third parameter specifying the mini-batch size, 
yielding a three-dimensional problem. Following the original authors, the search space is restricted to a
$6\times 6 \times 8$ grid~\citep{hoffman2010online}.
Next, we optimize a latent structured support vector machine (SVM) using a dataset available from~\cite{Snoek:2012}. Similarly to the LDA tuning experiment, the authors consider the latent structured
SVM on a three-dimensional parameter settings space of two regularisation parameters and a convergence tolerance. We used precomputed data corresponding to the
performance of the algorithm on binary classification of protein DNA
sequences.
Once again following the original methodology, the three-dimensional search space is discretised in a
$25~\times~14~\times~4$ grid~\citep{yu2009learning, miller2012max}.

Both datasets serve as benchmarks in the algorithm configuration 
community and the performance of a few different global optimisation approaches on these datasets is publicly available from~\cite{eggensperger2013towards}.

\begin {table*}[!t]
\begin{center}
\caption{HTBO performance evaluation}
\begin{tabular}{ r | r | r | r | r }
\label{tab:results}
{\bf Method}  &{\bf \textsc{BO}}  &{\bf \textsc{BO~Warp}}  &{\bf \textsc{HTBO}}  &{\bf \textsc{HTBO~Warp}}\\
	\hline
RKHS  & $-5.36 \pm 0.38$ & $-5.38 \pm 0.38$ & $-5.54 \pm 0.33$ & $\mathbf{-5.69} \pm 0.19$ \\
2-D Exp & $-0.24 \pm 0.20$ & $-0.14 \pm 0.18$ & $-0.39 \pm 0.11$ & $\mathbf{-0.41} \pm 0.07$ \\
LDA & $1266.26 \pm 0.30$ & $\mathbf{1266.16} \pm 0.00$ & $\mathbf{1266.16} \pm 0.00$ & $\mathbf{1266.16} \pm 0.00$ \\
SVM & $\mathbf{0.24} \pm 0.00$ & $\mathbf{0.24} \pm 0.00$ & $\mathbf{0.24} \pm 0.00$ & $\mathbf{0.24} \pm 0.00$  \\
Agromet & $-1050.09 \pm 100.69$ & $-1032.50 \pm 36.42$ & $\mathbf{-1117.39} \pm 32.82$ & $-1082.34 \pm 52.59$ \\
Brenda & $-0.19 \pm 0.07$ & $-0.21 \pm 0.12$ & $\mathbf{-0.26} \pm 0.12$ & $\mathbf{-0.26} \pm 0.09$
\end{tabular}
\end{center}
\caption*{The mean and standard deviation and the end of the scheduled iterations
with the best results in bold.
In most experiments, the proposed method
\textsc{HTBO~Warp} achieved the best performance, only with
\textsc{HTBO} achieving better results On Agromet.
\textsc{HTBO} also performed competitively in all experiments.} 
\end{table*}

\subsection{Geostatistical problems}
In geostatistics, Kriging is a method of interpolation with the aim
of modelling a function efficiently with a minimal number of observations.
Hence, Kriging is very closely related to Bayesian Optimisation.
In this subsection, we describe two datasets available at \url{kriging.com}. 
\begin{figure}[b!]
  \centering
  	\includegraphics[width=\columnwidth]{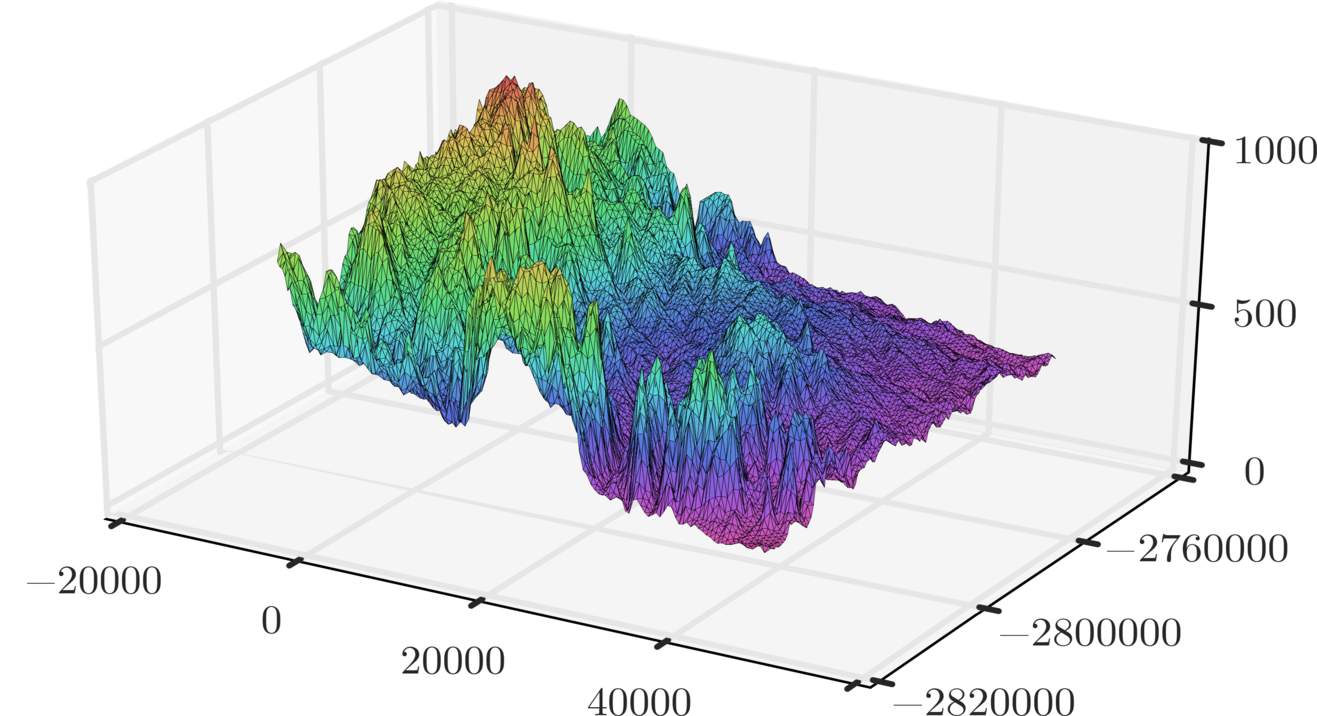}
	\caption{The surface described by the Agromet dataset. 
	The surface is heteroscedastic and rugged making it a difficult target for a single homoscedastic Gaussian process to model. }
 	\label{fig:func_agromet}
\end{figure}

The Agromet dataset was acquired by Isobel Clark, 
and describes a square area in the Natal Highlands, South Africa, 
measuring Gold grade of a drill hole intersection at $400$-meter spacing. 
Hence, this example is two-dimensional and is defined by the latitude 
and the longitude values of each acquired sample.
There are in total 18,189 observations.
As shown in~Figure~\ref{fig:func_agromet}, the surface described by the
dataset is clearly heteroscedastic and very rough,
making it difficult for a single Gaussian process, even with input warping, 
to model accurately. 

Brenda is an dataset of 1,856 observations of the depths of a
copper mine in British Columbia, Canada, that was closed
in 1990. Each observation in the dataset measures the concentrations of Copper, Molybdenum, Silver and Gold deposits. 
As in the case of Agromet, the surface described by Brenda also exhibits 
non-stationarity. 
In our experiments, we try only to identify the areas that contain the highest concentrations of Copper. 

Please refer to~\cite{clark2008practical} for more comprehensive descriptions
of these datasets.
In this paper, we use these two Kriging examples to construct optimisation problems 
by trying to find the highest concentrations of ores, based on the physically observed values of the datasets. Hence, we only query the expected improvement function at the points available from the historical records.

\subsection{Evaluation}
\label{sec:eval}
Each of the approaches was run $32$ times on all of the six benchmarks.
Figure~\ref{fig:results} summarises the average performance of all the runs.
In all examples, we try to minimize 
(instead of maximise) the objective function to follow the convention
of earlier work~\cite{Snoek:2012, snoek:2013b}.
We also report the mean and standard deviation of the runs in Table~\ref{tab:results}.

In the experiment involving the RKHS function (Figure~\ref{fig:results}~(a)), most of the \textsc{BO} and \textsc{BO~Warp} runs failed to converge to the global optimum. 
In contrast, by taking advantage of the tree partitioning, the proposed \textsc{HTBO} 
and \textsc{HTBO~Warp} approaches are capable of modelling 
and optimising the heteroscedastic objective function, with the latter achieving optimal performance.
Specifically, \textsc{HTBO~Warp} converged to the global optimum 
of the function in approximately $40$ evaluations.
As illustrated in Figure~\ref{fig:results}~(b), 
\textsc{BO~Warp} performs inadequately on the 2D Exp. function, while \textsc{HTBO} and \textsc{HTBO~Warp} exhibit the fastest convergence.
This illustrates the fact that input warping, despite being a very powerful technique, can fail in some heteroscedastic domains.

On the problem of online LDA, as shown in Figure~\ref{fig:results}~(c),
the best performance was achieved by \textsc{HTBO} and \textsc{HTBO~Warp}, with \textsc{BO~Warp} following closely.
On the structured SVM example, in Figure~\ref{fig:results}~(d), all methods have similar performance.
It appears that both problems are simple enough that all four methods
converge eventually. 
It is important to note that despite the lack of significant heteroscedasticity,
the proposed methods are still competitive with the state of art.

Finally, in the two real-world mining extraction problems \textsc{HTBO} and \textsc{HTBO Warp} achieve a significantly faster rate of convergence. They outperform  \textsc{BO} and \textsc{BO Warp}, that can't efficiently deal with the high heteroscedasticity of the domain. The performance is depicted in more detail in Figures~\ref{fig:results}~(e-f).

\section{Conclusion}
In this work, we introduced \textsc{HTBO}, a model based on decision trees with GP leaves, for tackling hard heteroscedastic functions in Bayesian optimisation. \textsc{HTBO} is a flexible model that does not place explicit requirements on its GP leaves and it can readily be combined with input warping (\textsc{HTBO Warp}).
We proposed a weighted marginal likelihood approach for learning the hyper-parameters of the leaves, and demonstrated empirically that our proposed methodological improvements have robust behaviour across a wide range of heteroscedastic functions. Finally, after evaluating the performance of \textsc{HTBO} and \textsc{HTBO Warp} on six problems, we showed that the proposed approaches outperform the competition in normal and heteroscedastic settings, and can yield both performance gains and robustness.

\subsubsection*{References}
\renewcommand\refname{\vskip -1.5em}
\setlength \bibhang{0in}
\bibliographystyle{include/natbib}

\bibliography{arXiv_treed_bo}

\end{document}

%% file: include/murphy.tex
\newcommand{\real}{\mathbb{R}}

\newcommand{\gauss}{{\cal N}}

\newcommand{\myvec}[1]{\mathbf{#1}}
\newcommand{\myvecsym}[1]{\boldsymbol{#1}}

\newcommand{\vtheta}{\myvecsym{\theta}}

\newcommand{\vk}{\myvec{k}}

\newcommand{\vm}{\myvec{m}}

\newcommand{\vx}{\myvec{x}}

\newcommand{\vy}{\myvec{y}}

\newcommand{\vI}{\myvec{I}}

\newcommand{\vK}{\myvec{K}}

\newcommand{\E}{\mathbb{E}}

\newcommand{\calD}{{\cal D}}

\newcommand{\calL}{{\cal L}}

\newcommand{\calX}{{\cal X}}

\newcommand{\data}{\calD}

\newcommand{\be}{\begin{equation}}
\newcommand{\ee}{\end{equation}}
\newcommand{\bea}{\begin{eqnarray}}
\newcommand{\eea}{\end{eqnarray}}
\newcommand{\beaa}{\begin{eqnarray*}}
\newcommand{\eeaa}{\end{eqnarray*}}

\DeclareMathAlphabet{\mathpzc}{OT1}{pzc}{m}{n}

